\documentclass[lettersize,journal]{IEEEtran}
\usepackage{amsmath,amssymb,amsfonts}
\usepackage{algorithmic}
\usepackage{algorithm}
\usepackage{array}
\usepackage{booktabs}
\usepackage{multirow}
\usepackage[caption=false,font=normalsize,labelfont=sf,textfont=sf]{subfig}
\usepackage{textcomp}
\usepackage{stfloats}
\usepackage{url}
\usepackage{verbatim}
\usepackage{graphicx}
\usepackage{cite}
\usepackage{xcolor}
\usepackage{makecell}
\usepackage{threeparttable}
\begin{document}

\title{A Distributed Multi-UGV Exploration Framework\\
With Loop-Aware Planning and Descriptor-Aided\\
Localization in Resource-Limited Environments}

\author{Zhiwei~Li,
        Haiou~Liu,
        Xijun~Zhao,
        Ji~Li,
        Yingze~Wang,
        and~Boyang~Wang%
\thanks{Received 10 September 2025; revised 3 January 2026 and 5 March 2026; accepted 7 April 2026. This work was supported by the National Natural Science Foundation of China under Grant 52302489 and Grant 52172378. Corresponding author: Boyang Wang.}%
\thanks{Zhiwei Li, Haiou Liu, Ji Li, and Yingze Wang are with the School of Mechanical Engineering, Beijing Institute of Technology, Beijing 100081, China (e-mail: 3220235285@bit.edu.cn; bit\_lho@bit.edu.cn; 3220240473@bit.edu.cn; 3220250382@bit.edu.cn).}%
\thanks{Xijun Zhao is with China North Artificial Intelligence \& Innovation Research Institute, Collective Intelligence \& Collaboration Laboratory (CIC), Beijing 100081, China (e-mail: 3220240470@bit.edu.cn).}%
\thanks{Boyang Wang is with the School of Mechanical Engineering, Beijing Institute of Technology, Beijing 100081, China, and also with Zhengzhou Intelligent Technology Research Institute, Beijing Institute of Technology, Zhengzhou 450046, China (e-mail: boyang\_wang@bit.edu.cn).}}

\markboth{IEEE Transactions on Industrial Electronics}%
{Li \MakeLowercase{\textit{et al.}}: A Distributed Multi-UGV Exploration Framework With Loop-Aware Planning and Descriptor-Aided Localization}

\maketitle

\begin{abstract}
Robust and efficient cooperative exploration with multiple unmanned ground vehicles (UGVs) in unknown, GPS-denied, and bandwidth-limited environments without prior maps remains challenging, as localization drift degrades map consistency and induces redundant coverage. This paper presents a fully distributed exploration framework that couples descriptor-aided inter-UGV loop closure with loop-aware hierarchical planning while enabling autonomous localization and exploration. We develop a lightweight LiDAR global descriptor with range-image prealignment to enable robust cross-UGV place recognition under large yaw and lateral variations, and use verified loop closures to maintain globally consistent trajectories and a sparse topological representation. We further introduce an uncertainty-aware cross-UGV loop-closure selection module that scores candidate loop closures under pose uncertainty and retains high-utility loop closures as planning anchors for global task allocation and local route refinement. Simulations and real-UGV experiments show that the loop-closure module achieves AR@1/AR@1\% of 89.9\%/95.5\%, distributed optimization reduces absolute trajectory error, the system substantially reduces two-way communication volume, and the overall framework reduces exploration time and travel distance by 15\% and 14\%, respectively, compared with an mTSP baseline.
\end{abstract}

\begin{IEEEkeywords}
Descriptor-aided localization, distributed SLAM, hierarchical exploration, loop-aware planning, multi-UGV systems, resource-limited environments.
\end{IEEEkeywords}

\section{Introduction}
\IEEEPARstart{C}{ollaborative} exploration and mapping using multi-UGV systems have attracted growing attention in applications such as disaster response, subterranean inspection, and planetary exploration. These missions are often conducted in resource-limited environments, characterized by GPS-denied conditions, limited communication bandwidth, and the absence of prior maps. In such settings, multi-UGV systems must perform accurate decentralized localization, consistent mapping, and efficient task coordination without relying on centralized infrastructure or global reference frames.

\begin{figure}[!t]
\centering
\includegraphics[width=\columnwidth]{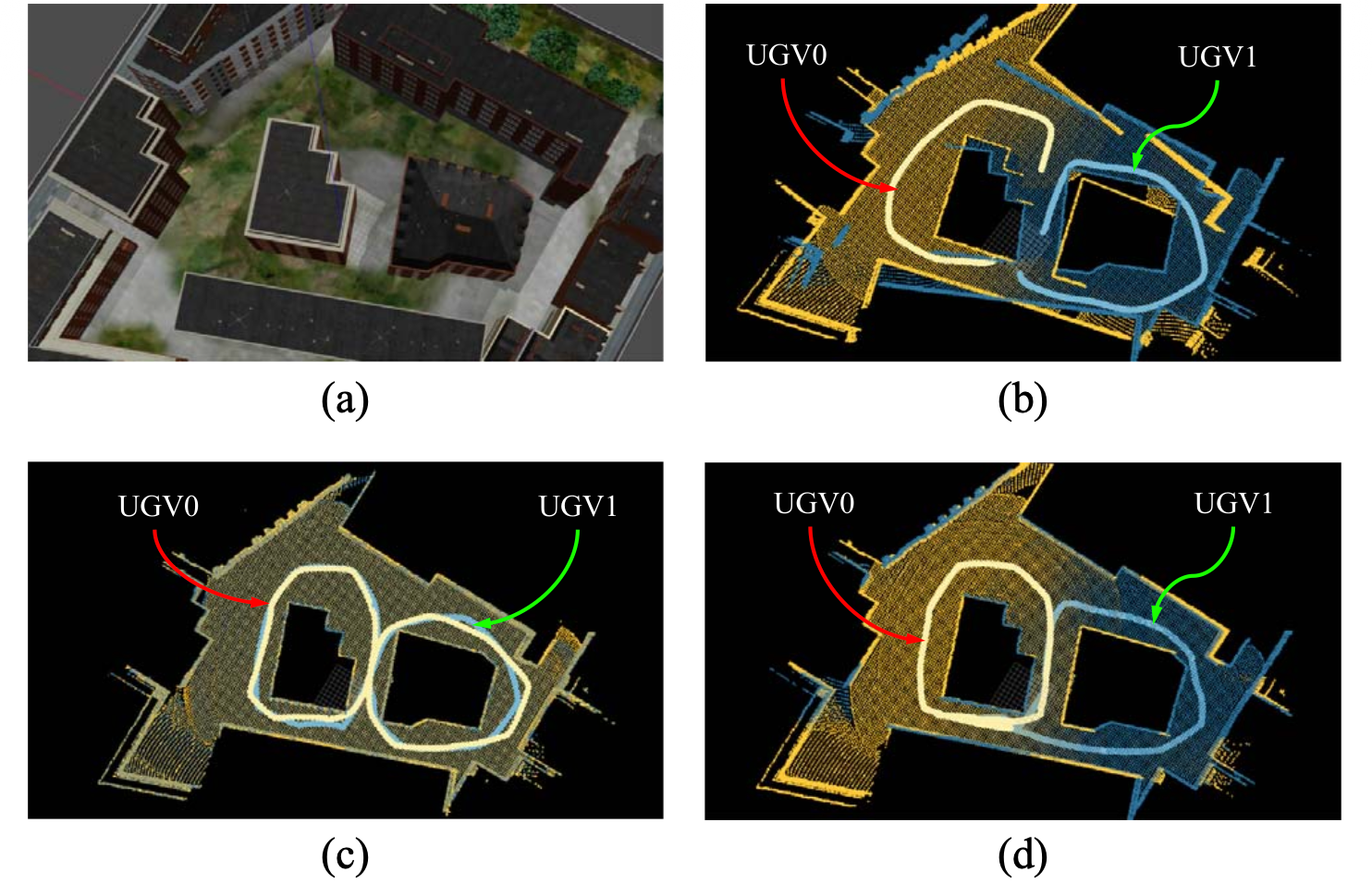}
\caption{Overview and motivation. Colored trajectories and point clouds show UGV paths and LiDAR observations. (a) Simulated urban scene with dense structures and narrow alleys. (b) GPS-denied exploration without reliable inter-UGV loop closures, causing drift and global misalignment. (c) Conservative revisits reduce drift but waste efficiency. (d) The proposed framework uses compact viewpoint-invariant descriptors and loop-aware planning to improve alignment and coordination.}
\label{fig:intro_motivation}
\end{figure}

A core challenge in distributed multi-UGV exploration is maintaining globally consistent localization across independent platforms despite severe viewpoint and lateral changes. In Fig.~\ref{fig:intro_motivation}(b), missing or unreliable cross-UGV loop closures in GPS-denied scenes accumulate pose drift, misalign maps, and degrade frontier detection. Fig.~\ref{fig:intro_motivation}(c) illustrates a conservative remedy: frequent revisits or large trajectory overlaps that curb drift but sharply reduce exploration efficiency through redundant coverage. These observations suggest that robust cross-UGV loop closures under significant yaw and lateral offsets are essential for achieving both accurate localization and efficient spatial coordination, as shown in Fig.~\ref{fig:intro_motivation}(d).

\begin{figure*}[!t]
\centering
\includegraphics[width=\textwidth]{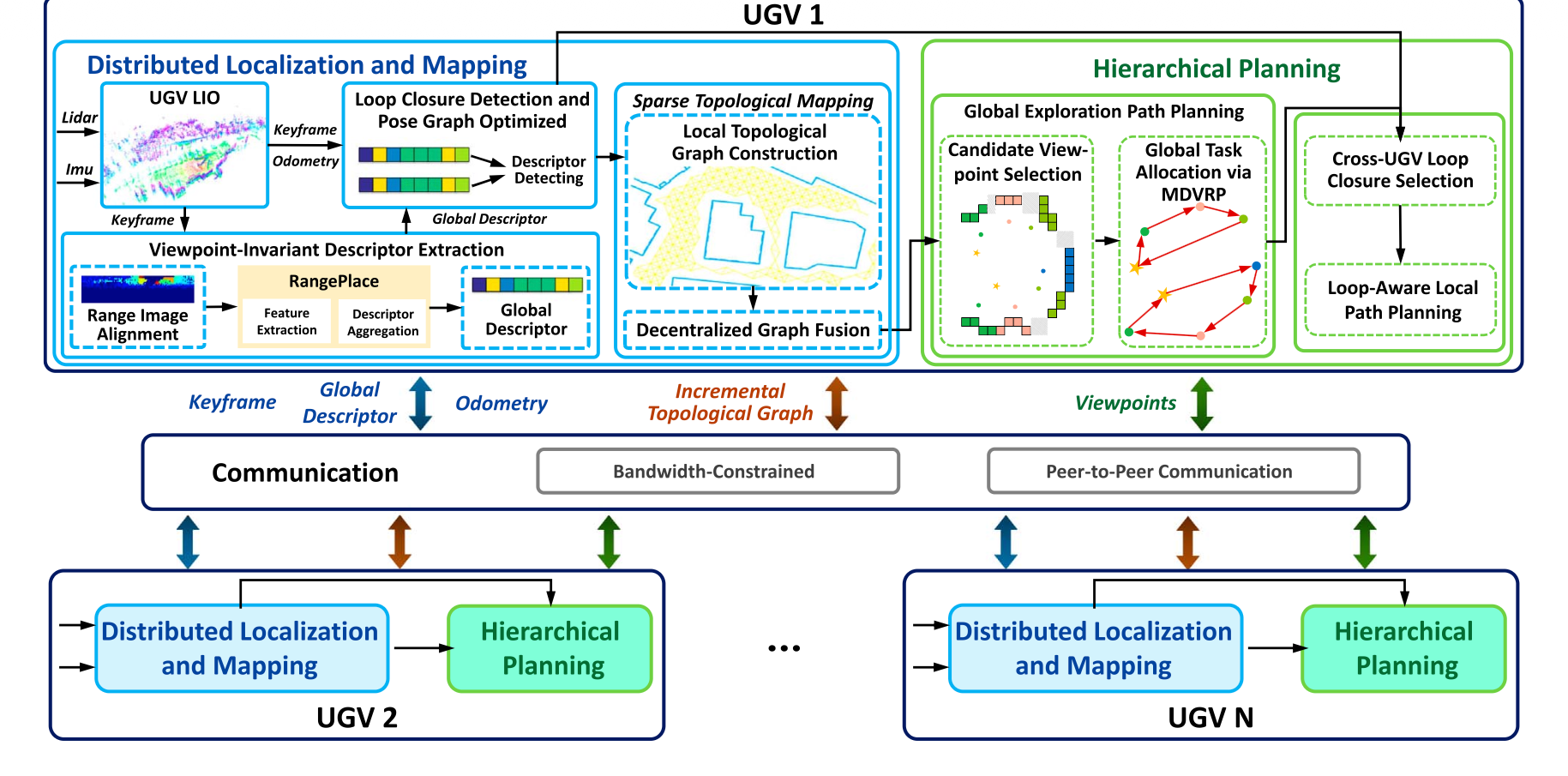}
\caption{Overview of the proposed collaborative exploration framework for multi-UGV systems in resource-limited environments. The architecture integrates distributed localization and mapping with loop-aware hierarchical planning. A distinctive feature is the proactive selection of cross-UGV loop closures as planning anchors guided by pose uncertainty, which enhances both global consistency and exploration efficiency.}
\label{fig:system_overview}
\end{figure*}

To address these challenges, we propose a fully distributed framework with two tightly coupled modules: 1) distributed localization and mapping, which uses a compact, viewpoint-invariant LiDAR descriptor with spectral-guided alignment for cross-UGV loop closure detection and maintains a sparse topological map with incremental, asynchronous subgraph fusion; and 2) loop-aware hierarchical planning, which scores inter-UGV loop closures using pose uncertainty, then inserts high-utility loop closures into local planning. Unlike prior systems~\cite{ref1,ref2,ref3,ref5,ref6} that treat loop closures as passive back-end constraints, our approach uses them as planning anchors, improving localization robustness, exploration efficiency, and communication economy. Fig.~\ref{fig:system_overview} illustrates the overall system architecture.

The main contributions of this work are summarized as follows.
\begin{enumerate}
\item We present a fully distributed multi-UGV exploration and mapping framework for GPS-denied, resource-limited, and prior-free environments. The framework tightly couples distributed localization, incremental mapping, and hierarchical planning, enabling reliable multi-UGV coordination and maintaining global map consistency under practical communication constraints.
\item We develop a compact LiDAR global descriptor with spectral-guided prealignment of range images, enabling robust cross-UGV place recognition and loop-candidate retrieval under large yaw changes and lateral shifts. Combined with a sparse topological representation and asynchronous subgraph fusion, the system shares descriptors and incremental subgraph updates rather than dense maps, substantially reducing communication overhead.
\item We introduce a loop-aware hierarchical exploration planning strategy that proactively selects high-utility inter-UGV loop opportunities via pose-uncertainty-driven evaluation, and incorporates the selected loop cues into both global task allocation and local path refinement to enhance localization robustness and exploration efficiency.
\end{enumerate}

The remainder of this article is structured as follows. Section~II reviews related work. Sections~III and IV describe the proposed framework in detail. Section~V presents experimental validation. Section~VI concludes the article.

\section{Related Work}
\subsection{Distributed SLAM and Loop Closure}
Collaborative simultaneous localization and mapping (C-SLAM) is vital to multi-UGV systems in GPS-denied, bandwidth-limited environments. Unlike centralized systems that depend on central servers or ultra-wideband anchors, distributed frameworks achieve scalability through local state estimation and selective information exchange. Yet most systems rely on line-of-sight inter-robot observations, which are hard to sustain in complex or large-scale environments.

Inter-robot loop closure detection remains a core challenge. To enable reliable inter-robot loop closure, recent systems adopt global descriptors, including rule-based methods like LiDAR-Iris, Scan Context++, and RING++, as well as learning-based approaches such as OverlapTransformer and RangePlace. While these improve viewpoint invariance, their robustness drops under severe lateral shifts or yaw discrepancies.

To ensure consistency, distributed pose graph optimization, such as distributed Gauss--Seidel and incremental solvers such as iSAM2, is widely used. Unlike sharing dense point clouds or voxel maps after optimization, which inflates bandwidth and latency, we keep a sparse topological graph of key poses and constraints and share only incremental subgraph updates. The descriptor-level decentralized workflow maintains global consistency and lightweight communication, making it suitable for long-duration missions.

\subsection{Exploration Planning and Task Allocation}
Recent advances improve exploration efficiency across trajectory generation, viewpoint selection, and map representation. TARE and representation granularity adopt hierarchical planning to balance global completeness with local feasibility, accelerating large-scale coverage. FAEP and uniform sampling reduce redundant maneuvers via frontier-level heuristics and hierarchical gain models. Lightweight GMM maps and topology-based abstractions further lower computation and bandwidth without sacrificing navigational safety.

For multirobot settings, decentralized coordination enhances task allocation. RACER and Varadharajan \emph{et al.} leverage topology-aware frontier assignment to balance load, while Zhao \emph{et al.} reduce redundancy through target selection and clustering. However, most existing frameworks treat loop closures as passive back-end constraints: they are incorporated only after being observed and are not evaluated by their heterogeneous utility in reducing pose uncertainty. Consequently, planners remain unaware of potentially informative loop opportunities, which can lead to drift accumulation or unnecessary trajectory overlap.

Recent efforts have started to exploit loop information more actively. Zhang \emph{et al.} couple exploration with Pose-SLAM uncertainty in a dual-layer planner, but focus on single-robot 2-D settings. Bai \emph{et al.} select informative loop edges using a log-determinant objective, yet assume a known topological map and centralized reasoning. Therefore, a decentralized, prior-free framework that explicitly scores loop-closure utility and integrates it into 3-D multirobot planning remains largely underexplored.

In contrast, we couple C-SLAM and exploration planning in a fully distributed stack. Inter-UGV loop closure candidates are scored by Lie-group uncertainty propagation and D-optimality. The most informative loop closures are integrated into both a multidepot vehicle routing problem (MDVRP) global allocator and a traveling salesman problem (TSP) local planner, thereby improving coverage, workload balance, and localization robustness in GPS-denied, low-bandwidth settings.

\section{Distributed Localization and Mapping}
This section describes the distributed localization and mapping module, which combines viewpoint-invariant loop closure detection, decentralized pose-graph optimization, and lightweight topological mapping. Each UGV extracts compact descriptors, detects cross-UGV loop closures, and incrementally exchanges local subgraphs to maintain global consistency under communication constraints.

\subsection{Viewpoint-Invariant Descriptor Extraction}
\subsubsection{Sensor Assumptions}
The proposed descriptor requires a rotating 3-D LiDAR with full $360^\circ$ horizontal FoV and multiple elevation channels, and is therefore not applicable to planar 2-D LiDARs. It is built from a multielevation range image and does not rely on a specific beam count or LiDAR model. The vertical FoV is unconstrained, except that vertical sampling must be dense enough to compute reliable range-image gradients. We used 16-beam and 64-beam LiDARs in our experiments. Configurations below 16 beams were not evaluated, while 16 beams and above should work.

\subsubsection{Spectral-Guided Saliency Alignment}
We build upon the original RangePlace by adding a lightweight prealignment module and retraining with diverse multi-UGV data. Unlike directly encoding raw range images, this canonicalizes the azimuthal bearing before descriptor extraction, improving cross-UGV consistency under yaw and lateral drift without changing the backbone or its inference complexity. We favor spectral and gradient saliency over entropy-like metrics since it more directly reflects stable geometric structure and is less sensitive to missing returns and dynamic clutter.

Given a scan, we form a spherical range image $R \in \mathbb{R}^{H \times W}$, where the row index corresponds to elevation and the column index corresponds to azimuth. For each azimuth column $c$, let $r_c(y)=R(y,c)$. A spectral cue is computed by summing the first $K_{\text{low}}$ low-frequency magnitudes of the 1-D discrete Fourier transform along the elevation direction:
\begin{equation}
E_c = \sum_{k=0}^{K_{\text{low}}-1} \left| \mathcal{F}\{r_c\}(k) \right|.
\label{eq:spectral}
\end{equation}

In parallel, a gradient cue measures range discontinuities using a robust median vertical difference:
\begin{equation}
G_c = \operatorname*{median}_{y \in \{0,\dots,H-2\}} \left|R(y+1,c)-R(y,c)\right|.
\label{eq:gradient}
\end{equation}

Both cues are smoothed along azimuth using a circular moving average:
\begin{equation}
\bar{E}_c = \frac{1}{w} \sum_{i=-\lfloor w/2 \rfloor}^{\lfloor w/2 \rfloor} E_{(c+i)\bmod W},
\quad
\bar{G}_c = \frac{1}{w} \sum_{i=-\lfloor w/2 \rfloor}^{\lfloor w/2 \rfloor} G_{(c+i)\bmod W}.
\label{eq:smooth}
\end{equation}

The saliency score is then defined as
\begin{equation}
S_c = \alpha \bar{E}_c + (1-\alpha)\bar{G}_c.
\label{eq:saliency}
\end{equation}
The canonical azimuth is chosen at the saliency maximum and the range image is circularly shifted accordingly:
\begin{equation}
\hat{R} = \left[ R(:,c^\star:W-1) \;\; R(:,0:c^\star-1) \right],
\quad
c^\star = \arg\max_c S_c.
\label{eq:canonical}
\end{equation}

The aligned range image is passed through a four-stage local Swin Transformer. Let $z_{l-1}$ denote the input to block $l$. The shifted-window attention and depthwise-convolution feed-forward steps are
\begin{align}
\hat{z}_l &= \operatorname{LN}\!\left(\operatorname{SW\mbox{-}MSA}(z_{l-1})\right) + z_{l-1}, \label{eq:swin1}\\
z_l &= \operatorname{LN}\!\left(\operatorname{DC\mbox{-}FFN}(\hat{z}_l)\right) + \hat{z}_l. \label{eq:swin2}
\end{align}
where $\hat{z}_l$ and $z_l$ denote the intermediate and final output features of block $l$, produced by the SW-MSA and DC-FFN modules, respectively. This process yields multiscale feature maps $\{F_1,F_2,F_3,F_4\}$, which are subsequently fused via a feature pyramid network and further refined by a multi-layer perceptron feature mix module. Finally, the fused representation is compressed through context gating to produce a compact 256-dimensional global descriptor.

Unlike RangePlace, which relies on learned feature invariance to implicitly handle rotational variations, our approach explicitly compensates for azimuthal misalignment before descriptor extraction. This lightweight preprocessing step improves descriptor consistency under large bearing changes without increasing network complexity, enabling high-recall loop closure detection with minimal communication overhead.

\subsection{Loop Closure Detection and Pose Graph Optimization}
Each UGV logs keyframe odometry and global descriptors, maintains them in a local KD-tree, and triggers peer-to-peer loop detection by exchanging descriptors upon new keyframes. Upon receiving a query descriptor, the nearest neighbor is retrieved, gated by descriptor similarity, and, if accepted, the corresponding keyframe point clouds are requested for Fast-GICP verification. Confirmed loop closures are incrementally inserted into a decentralized pose graph and optimized using iSAM2, enabling real-time updates with bounded computational cost.

All loop closure detection and pose graph optimization procedures are conducted asynchronously and locally. Only compact descriptors and verified proposals are communicated, minimizing bandwidth usage. Robustness is further enhanced through bidirectional matching and transformation consistency checks. This decentralized strategy ensures robust localization with globally consistent trajectories, even under sparse observations.

\subsection{Sparse Topological Mapping}
\begin{figure}[!t]
\centering
\includegraphics[width=\columnwidth]{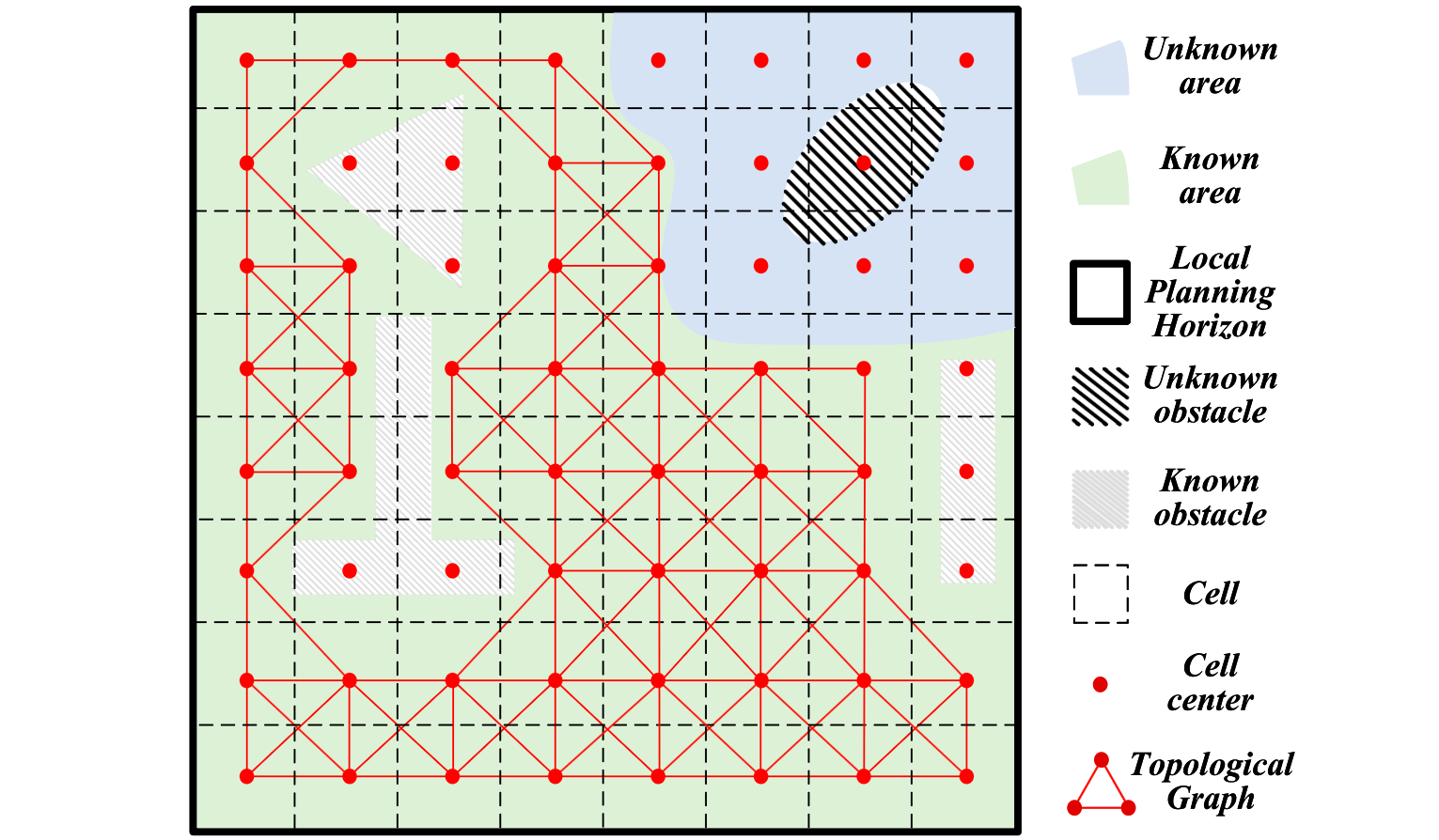}
\caption{Illustration of grid-based local topological graph construction.}
\label{fig:topology_graph}
\end{figure}

To represent the environment compactly for planning, we build a sparse, obstacle-aware topological graph within a local planning horizon of $L \times L$ centered at $R$. The space is voxelized and each voxel is labeled as traversable, occupied, or unknown from local point clouds. The centers of traversable voxels in $R$ form vertices and edges connect adjacent vertices when the straight-line segment is collision-free (Fig.~\ref{fig:topology_graph}). Only vertex coordinates and sparse adjacency are exchanged, reducing bandwidth while preserving obstacle constraints.

Subgraph exchange is driven by keyframes: when a new keyframe is created and optimized, the UGV extracts the incremental changes of the surrounding local topological graph, including updated vertices and edges, and transmits them as a subgraph for peer fusion via descriptor-aided node matching and edge merging. Algorithm~\ref{alg:fusion} outlines the fusion process, which preserves topological consistency while avoiding duplication. Collision checks are enforced before adding new edges to ensure path validity.

This decentralized and lightweight mapping strategy significantly reduces communication overhead and computation, while providing a globally consistent spatial structure for coordinated exploration. The resulting topological graph serves as a shared substrate for global task assignment and local path planning in the subsequent hierarchical planning module.

\begin{algorithm}[!t]
\caption{Decentralized Graph Fusion for Multi-UGV Collaboration}
\label{alg:fusion}
\begin{algorithmic}
\STATE \textbf{Input:} incoming subgraph $T_{\text{inc}}=(V_{\text{inc}},E_{\text{inc}})$, current graph $T=(V,E)$
\FORALL{$v \in V_{\text{inc}}$}
    \STATE find nearest node $v_{\text{match}}$ in $V$
    \IF{$v_{\text{match}}$ does not exist within merge radius}
        \STATE $V \leftarrow V \cup \{v\}$
    \ELSE
        \STATE merge the node identifiers of $v$ and $v_{\text{match}}$
    \ENDIF
\ENDFOR
\FORALL{$e \in E_{\text{inc}}$}
    \STATE obtain its endpoints $(v_i,v_j)$
    \IF{the edge does not exist and the segment is collision free}
        \STATE $E \leftarrow E \cup \{(v_i,v_j)\}$
    \ENDIF
\ENDFOR
\STATE \textbf{return} $T=(V,E)$
\end{algorithmic}
\end{algorithm}

\section{Hierarchical Planning}
This section describes the hierarchical planning module, which integrates global task allocation and local path refinement. Global exploration paths are assigned using an MDVRP-based allocator, while informative inter-UGV loop closures are evaluated and incorporated into local planning. This coupling enhances spatial coverage, workload balance, and localization consistency in distributed multi-UGV exploration.

\subsection{Global Exploration Path Planning}
\subsubsection{Frontier-Based Candidate Viewpoint Selection}
Each UGV begins by identifying frontier voxels defined as traversable cells adjacent to unexplored space. These frontiers are clustered based on spatial proximity to form potential exploration targets. For each cluster, candidate viewpoints are sampled along a circular shell centered at the cluster centroid to ensure directional diversity. To guarantee reachability, each candidate is projected onto its nearest node in the topological graph. A visibility-based scoring function is then used to evaluate the expected information gain of each candidate, and the one with the highest gain is selected per cluster. The resulting $V_{\text{task}}=\{v_1,\dots,v_M\}$ is used for global task assignment and exchanged at 1~Hz in our experiments.

\begin{figure}[!t]
\centering
\includegraphics[width=\columnwidth]{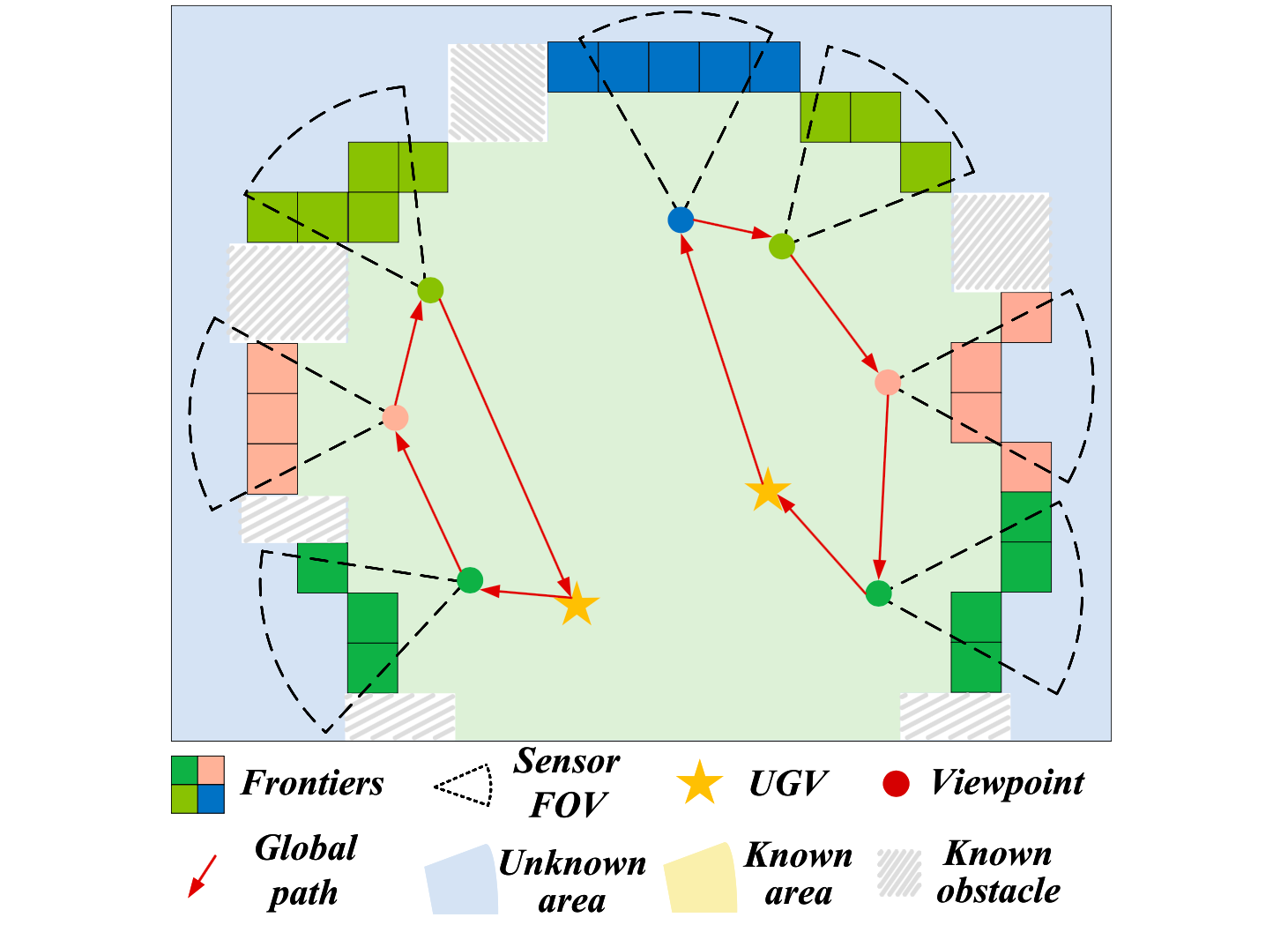}
\caption{Frontier-based multi-UGV exploration illustrating clustering, viewpoint selection, and path assignment.}
\label{fig:frontier_assignment}
\end{figure}

\subsubsection{Global Task Allocation via MDVRP}
Given the selected task viewpoints $V_{\text{task}}=\{v_1,\dots,v_M\}$, let $T=\{1,\dots,M\}$ denote the corresponding index set and $s_k$ denote the current node of UGV~$k$. We define the travel cost $c_{ij}$ between any two nodes $i,j \in T \cup \{s_k\}$ using A$^\ast$ search over the topological graph. Let $x_{ijk}\in\{0,1\}$ indicate whether UGV~$k$ travels from node $i$ to node $j$, and let $L_k$ be the route length of UGV~$k$. We introduce an auxiliary decision variable $L_{\max}$ to upper-bound all $L_k$ for workload balancing. The MDVRP is formulated as
\begin{equation}
\min \sum_{k=1}^{N} \sum_{i \in T \cup \{s_k\}} \sum_{j \in T \cup \{s_k\}} c_{ij} x_{ijk} + \lambda L_{\max}.
\label{eq:mdvrp_obj}
\end{equation}
The main constraints are
\begin{IEEEeqnarray}{rCl}
\sum_{k=1}^{N} \sum_{j \in T \cup \{s_k\}} x_{ijk}
&=&
1, \quad \forall i \in T
\label{eq:mdvrp_assign}\\
\sum_{j \in T} x_{s_kjk}
&=&
1, \quad \forall k \in \{1,\dots,N\}
\label{eq:mdvrp_depart}\\
\sum_{i \in T \cup \{s_k\}} x_{is_kk}
&=&
0, \quad \forall k \in \{1,\dots,N\}
\label{eq:mdvrp_open}\\
\sum_{i \in T \cup \{s_k\}} x_{ihk}
&=&
\sum_{j \in T \cup \{s_k\}} x_{hjk},
\quad \forall h \in T, \forall k
\label{eq:mdvrp_flow}\\
L_k
&=&
\sum_{i \in T \cup \{s_k\}} \sum_{j \in T \cup \{s_k\}} c_{ij} x_{ijk},
\quad \forall k
\label{eq:mdvrp_length}\\
L_k
&\leq&
L_{\max}, \quad \forall k.
\label{eq:mdvrp_balance}
\end{IEEEeqnarray}

Constraints~\eqref{eq:mdvrp_assign}--\eqref{eq:mdvrp_depart} assign each viewpoint exactly once and enforce one departure from each depot. Constraints~\eqref{eq:mdvrp_open}--\eqref{eq:mdvrp_flow} impose flow conservation and allow an open route for continuous exploration. Constraints~\eqref{eq:mdvrp_length}--\eqref{eq:mdvrp_balance} define $L_k$ and bound it by $L_{\max}$ for workload balance, weighted by $\lambda>0$.

\subsection{Cross-UGV Loop Closure Selection}
Notations. $\operatorname{Exp}(\cdot):\mathfrak{se}(3)\rightarrow SE(3)$ and $\operatorname{Log}(\cdot):SE(3)\rightarrow \mathfrak{se}(3)$ denote the exponential and logarithm maps. $[\cdot]^\wedge:\mathbb{R}^6\rightarrow\mathfrak{se}(3)$ is the wedge operator, and $(\cdot)^\vee$ is its inverse. $\operatorname{Ad}(\cdot)$ denotes the adjoint operator on $SE(3)$. $\operatorname{trans}(\cdot)$ extracts the translation component of an $SE(3)$ pose.

After global task allocation, UGV~$p$ propagates open-loop uncertainty along its planned segment.

For UGV $p$, let $\bar{X}^{p}_{n}$ denote the predicted mean pose at step $n$ and let $\Omega^p_n$ denote the nominal increment from step $n-1$ to $n$. Mean propagation is written as
\begin{equation}
\bar{X}^{p}_{n} = \bar{X}^{p}_{n-1}\operatorname{Exp}\!\left([\Omega^p_n]^\wedge\right).
\label{eq:meanprop}
\end{equation}
The actual motion is modeled as
\begin{equation}
X^p_n = X^p_{n-1}\operatorname{Exp}\!\left([\Omega^p_n+w^p_n]^\wedge\right),
\label{eq:motion}
\end{equation}
where $w^p_n \sim \mathcal{N}(0,Q^p_n)$. Using a right-invariant error model,
\begin{equation}
X^p_n = \bar{X}^p_n \operatorname{Exp}\!\left([\xi^p_n]^\wedge\right),
\quad
\xi^p_n \sim \mathcal{N}(0,\Sigma^p_{n,n}),
\label{eq:riem_error}
\end{equation}
the covariance propagation becomes
\begin{equation}
\Sigma^p_{n,n} = F^p_n \Sigma^p_{n-1,n-1} (F^p_n)^\top
+ \Phi(\Omega^p_n)Q^p_n\Phi(\Omega^p_n)^\top.
\label{eq:covprop}
\end{equation}

Let $R_p$ denote the local planning region of UGV~$p$, let $\mathcal{L}_p$ denote the index set of sampled predicted poses on the planned segment inside $R_p$, and let $M_{pq}$ denote the keyframes of another UGV~$q$ that fall inside the same region. For each candidate pair $(l,m)$, a prediction-time relative-pose measurement model is used to define an innovation covariance
\begin{equation}
S^{pq}_{lm,\text{pred}} = H^{pq}_{lm}\Sigma^{pq}_{lm}(H^{pq}_{lm})^\top + R_{lm,\text{pred}}^{pq}.
\label{eq:innovation_cov}
\end{equation}
The utility score for ranking candidate loop closures is
\begin{equation}
u^{pq}_{lm,\text{pred}} =
\left[\log\det\!\left(S^{pq}_{lm,\text{pred}}\right)-\log\det\!\left(R^{pq}_{lm,\text{pred}}\right)\right]
\exp(-\gamma_d d^{pq}_{lm}),
\label{eq:utility}
\end{equation}
where $d^{pq}_{lm}$ is the Euclidean distance between the predicted and historical positions and $\gamma_d>0$ is a distance-decay coefficient.

Algorithm~\ref{alg:selection} summarizes the selection procedure. The top-$K$ ranked candidates are retained as planning anchors.

\begin{algorithm}[!t]
\caption{Cross-UGV Loop Closure Selection}
\label{alg:selection}
\begin{algorithmic}
\STATE \textbf{Input:} predicted states in local region $R_p$, global path $P^p_g$, historical keyframes from peer UGVs, cooldown gate, top-$K$
\STATE compute intersections between $P^p_g$ and $\partial R_p$
\STATE sample predicted poses along the local segment of $P^p_g$
\FORALL{peer UGVs $q \neq p$}
    \FORALL{peer keyframes inside $R_p$}
        \STATE associate each peer keyframe with the nearest predicted pose
        \IF{the cooldown gate is open}
            \STATE add the pair to the candidate set
        \ENDIF
    \ENDFOR
\ENDFOR
\FORALL{candidate pairs}
    \STATE compute the predicted utility score using \eqref{eq:utility}
\ENDFOR
\STATE rank candidates by utility and keep the top-$K$
\STATE \textbf{return} the selected loop-candidate set and the local intersections
\end{algorithmic}
\end{algorithm}

\subsection{Loop-Aware Local Path Planning}
Within the local planning region $R_p$, the selected loop candidates are converted into loop-aware waypoints by projecting peer keyframe positions onto the current topological graph:
\begin{equation}
V^p_{\text{loop}} =
\left\{
\Pi_{\mathcal{T}}\!\left(\operatorname{trans}(\hat{X}^q_m)\right)
\; \middle| \; (q,l,m)\in C^p_{\text{loop}}
\right\}.
\label{eq:loop_wp}
\end{equation}
The informative waypoint set is then defined as
\begin{equation}
V^p_{\text{info}} = I_p \cup V^p_{\text{frontier}} \cup V^p_{\text{loop}},
\label{eq:info_wp}
\end{equation}
where $I_p$ are boundary intersections and $V^p_{\text{frontier}}$ are frontier viewpoints inside $R_p$. A symmetric TSP is finally solved on $V^p_{\text{info}}$ using edge costs computed by shortest-path search on the topological graph. This produces a loop-aware local plan that favors informative revisits while suppressing unnecessary overlap.

\section{Experiments}
In this section, we validate our framework through both simulation and real-world experiments, focusing on localization accuracy, exploration performance, and communication cost. Consistent sensor and computing setups are used across all platforms to ensure fair comparison.

\subsection{Experimental Platform}
\begin{figure}[!t]
\centering
\includegraphics[width=\columnwidth]{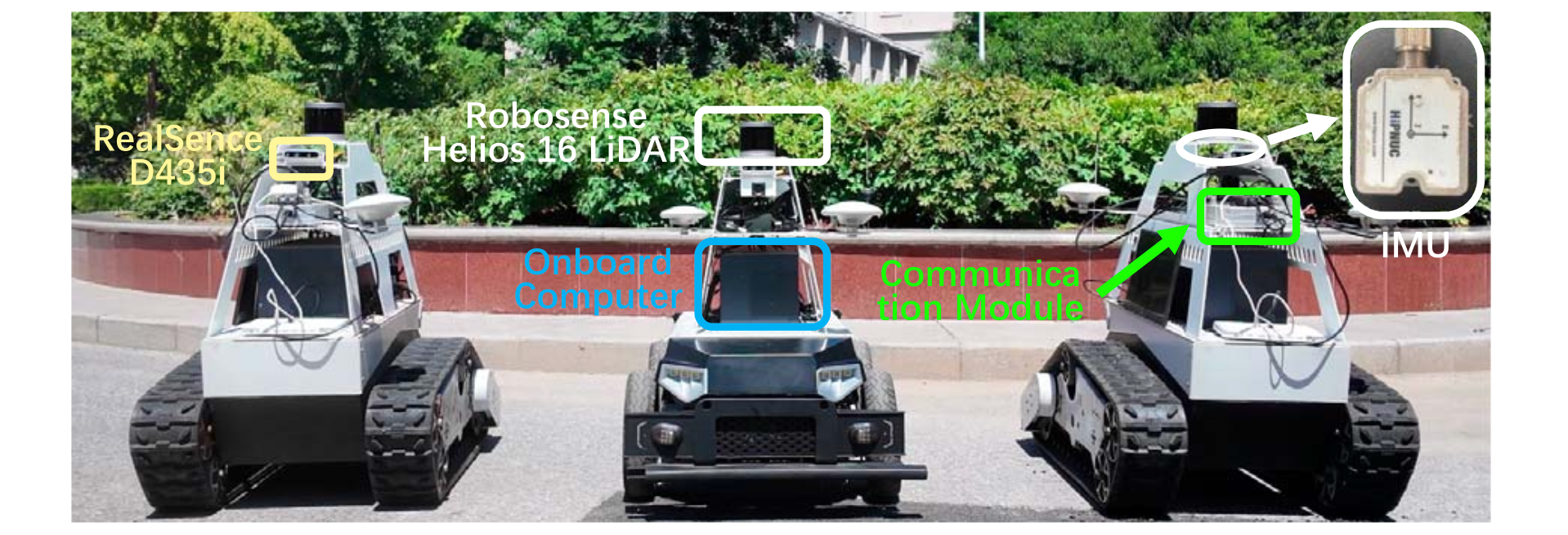}
\caption{UGV platforms used in the experiments. Each UGV is equipped with LiDAR, IMU, onboard computation, and a communication module to support decentralized mapping and exploration.}
\label{fig:platform}
\end{figure}

We evaluate the proposed system through both real-world tests and Gazebo-based simulations. As shown in Fig.~\ref{fig:platform}, each UGV is equipped with a RoboSense Helios-16 LiDAR, a Hipnuc IMU, a RealSense D435i depth camera, and an onboard computer (Intel i9-14900K CPU, 32~GB DDR4 RAM, and Nvidia RTX~4080 GPU). Inter-UGV communication is implemented via a local area network, and the simulation environment replicates the same hardware and sensor configuration to ensure consistency.

\subsection{Evaluation of Distributed Localization and Mapping}
\noindent This section presents comprehensive evaluations of our distributed localization method, which leverages global descriptors for inter-UGV loop closure detection. Our method is designed to be descriptor-agnostic, allowing the integration of different global descriptors in a modular fashion.

\subsubsection{Cross-UGV Loop Closure Detection}
To evaluate descriptor robustness under diverse real-world conditions, we select five representative sequences from the KITTI and Mulran datasets, each representing a distinct viewpoint challenge: KITTI~00 (dynamic occlusion with moderate lateral shift), KITTI~02 (significant yaw variation), KITTI~05 (stable short-range path), KITTI~08 (reverse revisits with lateral drift), and Mulran Riverside~02 (same-direction revisits with wide-lane displacement). We evaluate three global descriptors integrated into our distributed loop closure detection framework: Scan Context++, OverlapTransformer, and our proposed method. In addition, we include LiDAR-Iris, as used in DCL-SLAM, for baseline comparison. Table~\ref{tab:descriptor} reports their Average Recall at Top-1 (AR@1) and Top-1\% (AR@1\%).

\begin{table*}[!t]
\caption{AR@1 and AR@1\% Performance of Different Global Descriptors Across Five Benchmark Scenes}
\label{tab:descriptor}
\centering
\resizebox{\textwidth}{!}{%
\begin{tabular}{lcccccccccccc}
\toprule
\multirow{2}{*}{Method} & \multicolumn{2}{c}{KITTI 00} & \multicolumn{2}{c}{KITTI 02} & \multicolumn{2}{c}{KITTI 05} & \multicolumn{2}{c}{KITTI 08} & \multicolumn{2}{c}{Mulran R. 02} & \multicolumn{2}{c}{Mean Average} \\
& AR@1 & AR@1\% & AR@1 & AR@1\% & AR@1 & AR@1\% & AR@1 & AR@1\% & AR@1 & AR@1\% & AR@1 & AR@1\% \\
\midrule
L-Iris~\cite{ref13}            & 92.63 & 96.02 & 78.77 & 85.71 & 87.22 & 91.28 & 83.46 & 93.12 & 80.75 & 88.79 & 84.57 & 90.98 \\
SC++~\cite{ref14}              & 82.18 & 85.26 & 71.30 & 75.43 & 75.68 & 82.80 & 66.03 & 76.38 & 66.88 & 92.30 & 72.41 & 82.43 \\
OT~\cite{ref16}                & 93.85 & 97.93 & 85.71 & 93.33 & 83.91 & 90.04 & 67.76 & 89.31 & 72.14 & 92.75 & 80.67 & 92.67 \\
Ours                           & 96.29 & 98.62 & 95.73 & 97.60 & 90.05 & 92.01 & 86.21 & 95.34 & 81.20 & 93.99 & 89.90 & 95.51 \\
\bottomrule
\end{tabular}}
\end{table*}

As shown in Table~\ref{tab:descriptor}, our descriptor consistently outperforms all baselines, especially in sequences with significant viewpoint variations. Scan Context++ exhibits performance degradation under yaw rotation and reverse revisits, while OverlapTransformer is sensitive to lateral drift. LiDAR-Iris performs reasonably well overall but shows reduced stability in more challenging scenarios.

\subsubsection{Distributed Localization Performance}
To assess distributed localization performance, we evaluate Absolute Trajectory Error (ATE) across four representative scenarios: two from the S3E benchmark and two from real-world deployments. These cover a range of structural and environmental characteristics: Library (S3E) with dense structures and frequent loop closures, Playground (S3E) as an open space with sparse features, Scene~1 (ours) as a medium-scale industrial park with appearance aliasing and short overlapping paths, and Scene~2 (ours) as a large suburban area with similar aliasing but limited loop closure opportunities. We compare seven configurations: Fast-LIO2 (standalone), DCL-SLAM (with LiDAR-Iris), DiSCo-SLAM, Kimera-Multi, and our system using three different descriptors. Results are reported in Table~\ref{tab:ate}.

\begin{figure}[!t]
\centering
\includegraphics[width=\columnwidth]{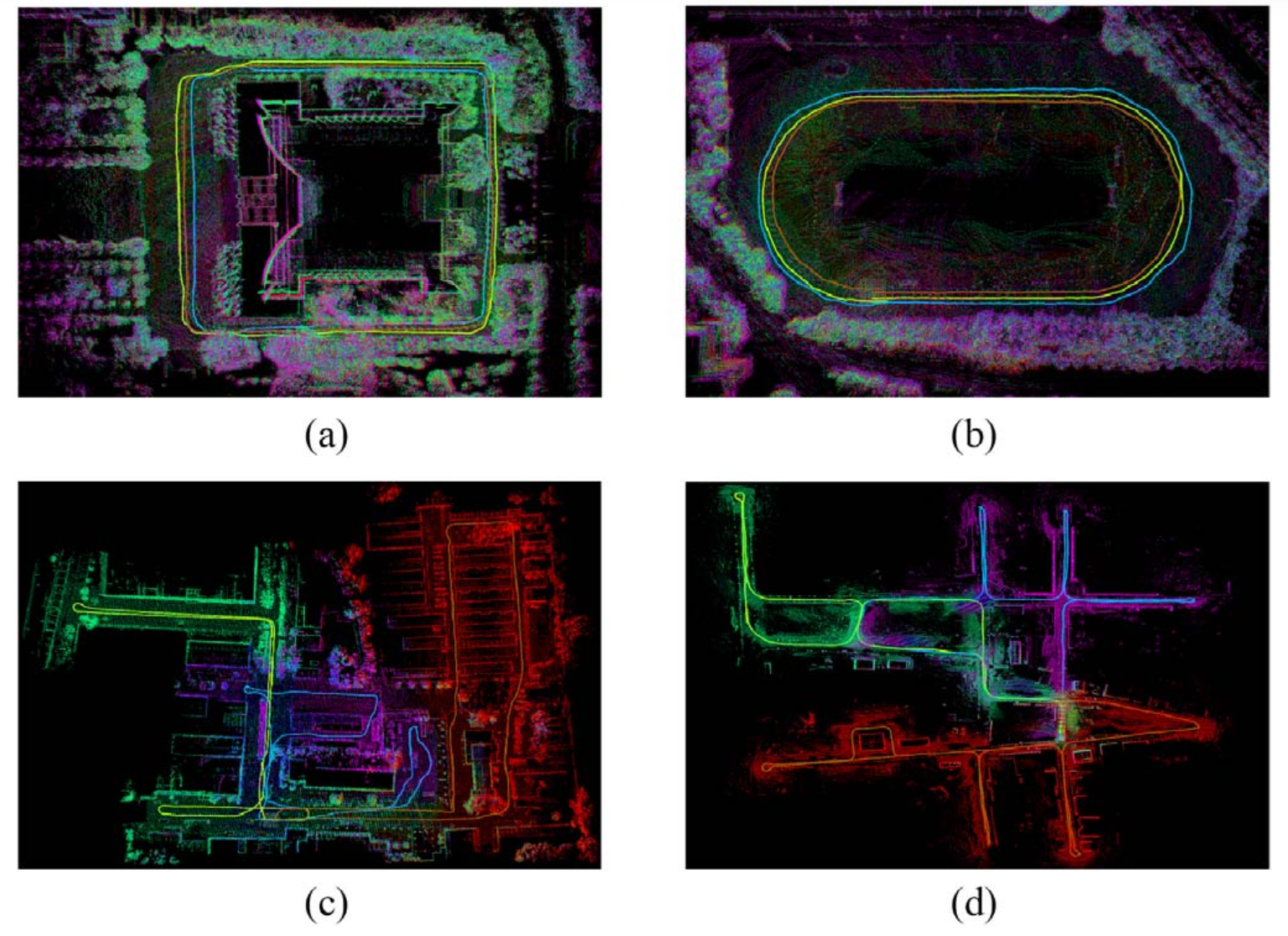}
\caption{Visualization of mapping results across four representative environments. Red, green, and blue point clouds denote data from three UGVs, with trajectories in matching colors. The scenarios cover loop-rich, sparse-feature, aliasing-prone, and loop-scarce environments, jointly validating accuracy, robustness, and adaptability.}
\label{fig:mapping_results}
\end{figure}

\begin{table*}[!t]
\caption{ATE (m) Comparison Across Four Scenarios}
\label{tab:ate}
\centering
\resizebox{\textwidth}{!}{%
\begin{tabular}{lcccccccccccc}
\toprule
\multirow{2}{*}{Method} & \multicolumn{3}{c}{Library} & \multicolumn{3}{c}{Playground} & \multicolumn{3}{c}{Scene 1} & \multicolumn{3}{c}{Scene 2} \\
& UGV1 & UGV2 & UGV3 & UGV1 & UGV2 & UGV3 & UGV1 & UGV2 & UGV3 & UGV1 & UGV2 & UGV3 \\
\midrule
Only Fast-LIO2~\cite{ref29}            & 1.468 & 2.469 & 1.389 & 0.659 & 1.518 & 1.019 & 2.288 & 1.856 & 1.165 & 6.830 & 3.734 & 3.503 \\
DCL-SLAM (Lidar-Iris~\cite{ref13})     & 1.364 & 1.823 & 1.459 & 2.286 & 1.437 & 1.315 & 7.542 & 27.828 & 5.693 & 7.712 & 5.473 & 3.817 \\
DiSCo-SLAM~\cite{ref10}                & 5.497 & 0.860 & 1.163 & 0.607 & 1.187 & 0.982 & 1.119 & 1.032 & 1.074 & 6.120 & 1.804 & 1.633 \\
Kimera-Multi~\cite{ref30}              & -- & -- & -- & 3.963 & 4.251 & 4.893 & 10.052 & 9.763 & 9.385 & -- & -- & -- \\
Ours (Scan Context++~\cite{ref14})     & 1.925 & 1.809 & 1.791 & 0.723 & 1.092 & 1.087 & 1.126 & 1.038 & 1.067 & 5.353 & 1.803 & 1.634 \\
Ours (OverlapTransformer~\cite{ref16}) & 1.454 & 1.570 & 1.343 & 0.841 & 1.111 & 1.087 & 1.098 & 1.020 & 1.020 & 5.352 & 1.805 & 1.642 \\
Ours                                   & 0.449 & 1.069 & 0.847 & 0.423 & 0.761 & 0.795 & 0.887 & 0.816 & 0.882 & 4.155 & 1.961 & 1.222 \\
\bottomrule
\end{tabular}}
\end{table*}

\subsubsection{Robustness Under Network Impairments}
Latency and bandwidth stress tests are conducted by injecting communication delay or limiting application-layer goodput. Table~\ref{tab:delay} reports ATE and communication rate under delay; Table~\ref{tab:bandwidth} reports ATE under bandwidth caps. The system remains stable under moderate impairment and degrades more noticeably when delays exceed 3~s or the bandwidth cap drops to 0.6~Mbps. This behavior matches the intuition that delayed loop constraints and slower keyframe propagation weaken drift correction.

\begin{table*}[!t]
\caption{ATE (m) and Average Communication Rate Under Different Communication Delays}
\label{tab:delay}
\centering
\resizebox{\textwidth}{!}{%
\begin{tabular}{lcccccccccccccccc}
\toprule
\multirow{2}{*}{Delay (s)} & \multicolumn{4}{c}{Library} & \multicolumn{4}{c}{Playground} & \multicolumn{4}{c}{Scene 1} & \multicolumn{4}{c}{Scene 2} \\
& UGV1 & UGV2 & UGV3 & KB/s & UGV1 & UGV2 & UGV3 & KB/s & UGV1 & UGV2 & UGV3 & KB/s & UGV1 & UGV2 & UGV3 & KB/s \\
\midrule
0.5 $\pm$ 0.1 & 0.449 & 1.057 & 0.823 & 49.98 & 0.399 & 0.809 & 0.823 & 96.90 & 0.887 & 0.854 & 0.888 & 27.42 & 3.299 & 4.276 & 3.592 & 54.66 \\
1.0 $\pm$ 0.1 & 0.449 & 1.088 & 0.833 & 98.26 & 0.399 & 0.812 & 0.771 & 74.78 & 0.887 & 0.879 & 0.999 & 59.01 & 3.299 & 4.097 & 3.832 & 53.16 \\
2.0 $\pm$ 0.1 & 0.451 & 1.168 & 1.165 & 60.07 & 0.399 & 1.006 & 0.907 & 93.85 & 0.887 & 0.958 & 0.948 & 56.24 & 3.479 & 4.035 & 3.652 & 99.63 \\
3.0 $\pm$ 0.1 & 0.449 & 2.061 & 1.368 & 45.05 & 0.399 & 1.487 & 3.148 & 64.97 & 2.063 & 3.667 & 1.351 & 25.07 & 6.729 & 8.321 & 7.526 & 90.22 \\
5.0 $\pm$ 0.1 & 0.449 & 10.687 & 4.869 & 39.82 & 0.399 & 5.129 & 1.783 & 104.84 & 2.063 & 3.667 & 4.153 & 23.12 & 6.929 & 9.798 & 7.165 & 54.20 \\
\bottomrule
\end{tabular}}
\end{table*}

\begin{table}[!t]
\caption{ATE (m) Under Different Bandwidth Caps (No Extra Delay)}
\label{tab:bandwidth}
\centering
\resizebox{\columnwidth}{!}{%
\begin{tabular}{lcccc}
\toprule
Environment & Bandwidth (Mbps) & UGV1 & UGV2 & UGV3 \\
\midrule
\multirow{3}{*}{Library}    & 2.5 & 0.449 & 1.057 & 0.823 \\
                            & 1.2 & 0.447 & 1.086 & 1.117 \\
                            & 0.6 & 0.447 & 2.679 & 2.872 \\
\multirow{3}{*}{Playground} & 2.5 & 0.423 & 0.761 & 0.795 \\
                            & 1.2 & 0.399 & 0.862 & 1.620 \\
                            & 0.6 & 0.399 & 1.751 & 4.023 \\
\multirow{3}{*}{Scene 1}    & 2.5 & 0.887 & 0.820 & 0.882 \\
                            & 1.2 & 0.887 & 1.091 & 1.293 \\
                            & 0.6 & 0.887 & 3.969 & 3.784 \\
\multirow{3}{*}{Scene 2}    & 2.5 & 3.297 & 3.415 & 3.358 \\
                            & 1.2 & 3.299 & 3.414 & 3.168 \\
                            & 0.6 & 7.010 & 3.409 & 4.172 \\
\bottomrule
\end{tabular}}
\end{table}

\subsubsection{Scalability of Communication and Computation}
To study scalability, the team size is increased from three to six UGVs. Table~\ref{tab:scale} shows that both communication and computation grow moderately. The main communication cost comes from loop-related keyframe payloads, whereas odometry and descriptor exchange remain comparatively lightweight. Descriptor generation, local optimization, and KD-tree retrieval remain fast, while global optimization grows the most due to additional keyframes and inter-robot constraints.

\begin{table*}[!t]
\caption{Scalability of Communication and Computation With Different Numbers of UGVs}
\label{tab:scale}
\centering
\resizebox{\textwidth}{!}{%
\begin{tabular}{cccccccccc}
\toprule
\multirow{2}{*}{UGVs} & \multicolumn{5}{c}{Communication Cost (Aggregate)} & \multicolumn{4}{c}{Computation Time} \\
\cmidrule(lr){2-6}\cmidrule(lr){7-10}
& Odom. (MB) & PR (MB) & PC (MB) & Total (MB) & KB/s & PR Gen. (ms) & KD-tree + Query (ms) & Local Opt. (ms) & Global Opt. (s) \\
\midrule
3 & 0.21 & 3.88 & 78.80 & 82.89 & 85.74 & 12.87 $\pm$ 4.95 & 1.58 $\pm$ 5.31 & 2.35 $\pm$ 2.02 & 2.40 $\pm$ 2.59 \\
4 & 0.34 & 6.14 & 85.37 & 91.84 & 95.00 & 13.83 $\pm$ 4.73 & 1.60 $\pm$ 5.62 & 2.81 $\pm$ 2.54 & 2.59 $\pm$ 2.49 \\
5 & 0.48 & 8.85 & 89.12 & 98.45 & 101.83 & 13.95 $\pm$ 5.58 & 1.48 $\pm$ 7.08 & 2.50 $\pm$ 1.80 & 3.50 $\pm$ 3.12 \\
6 & 0.67 & 12.23 & 95.70 & 108.60 & 112.32 & 16.25 $\pm$ 8.91 & 2.82 $\pm$ 21.54 & 2.89 $\pm$ 3.20 & 4.12 $\pm$ 4.31 \\
\bottomrule
\end{tabular}}
\end{table*}

\subsection{Hierarchical Planning Ablation Study}
To isolate the benefit of loop-aware planning, the full method (MDVRP+LOOP) is compared with three variants: mTSP, mTSP+LOOP, and MDVRP without loop-aware local refinement. All methods share the same planning and control modules apart from the task-allocation and loop-integration differences.

\begin{table*}[!t]
\caption{Performance Comparison of Different Hierarchical Planning Strategies}
\label{tab:planning}
\centering
\resizebox{\textwidth}{!}{%
\begin{tabular}{lcccccccccc}
\toprule
\multirow{2}{*}{Method} & \multicolumn{2}{c}{Exploration Time (s)} & \multicolumn{2}{c}{Total Distance (m)} & \multicolumn{2}{c}{Loop Closures} & \multicolumn{2}{c}{Overlap Distance (m)} & \multicolumn{2}{c}{Path Balance} \\
& Mean & Std & Mean & Std & Mean & Std & Mean & Std & Mean & Std \\
\midrule
MTSP~\cite{ref1}         & 305.29 & 42.03 & 1683.95 & 236.69 & 419.50 & 154.79 & 532.39 & 229.29 & 0.976 & 0.00028 \\
MTSP~\cite{ref1}+LOOP    & 285.80 & 26.39 & 1580.24 & 143.06 & 406.20 & 143.44 & 444.05 & 121.82 & 0.984 & 0.00003 \\
MDVRP                    & 278.09 & 42.04 & 1565.43 & 171.13 & 365.70 & 174.73 & 425.44 & 172.10 & 0.977 & 0.00023 \\
MDVRP+LOOP               & 258.10 & 30.18 & 1443.81 & 162.13 & 217.30 & 70.48  & 259.99 & 112.77 & 0.985 & 0.00003 \\
\bottomrule
\end{tabular}}
\end{table*}

The proposed MDVRP+LOOP strategy yields the best overall performance, reducing exploration time and travel distance by 15.4\% and 14.3\% relative to the mTSP baseline. It also produces fewer loop closures and over 50\% less overlap distance, indicating more targeted revisits and improved spatial coordination.

\begin{figure}[!t]
\centering
\includegraphics[width=\columnwidth]{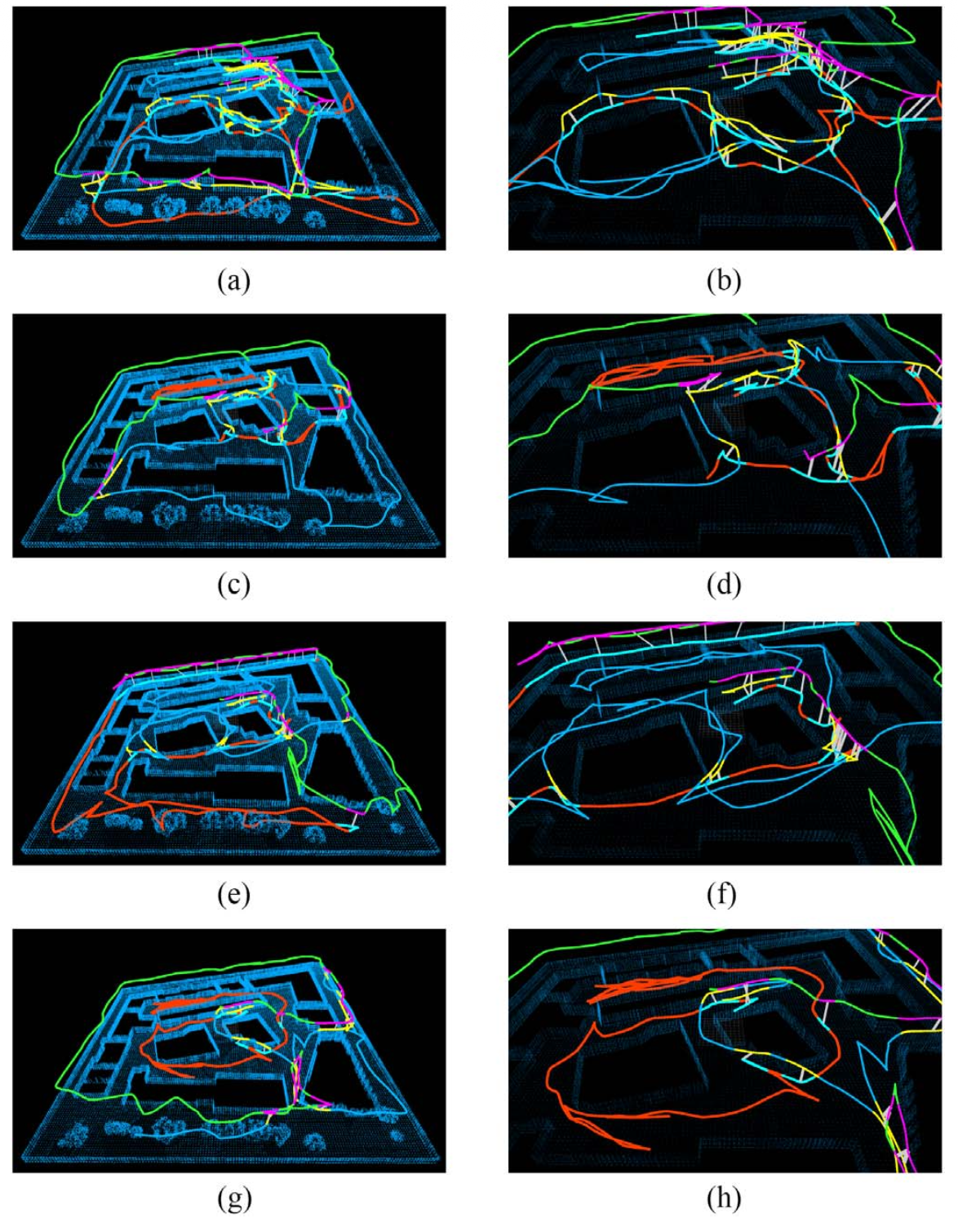}
\caption{Qualitative comparison of multi-UGV exploration trajectories under different hierarchical planning strategies. The left column shows complete exploration paths, while the right column provides a zoomed-in view of the central area to highlight trajectory interactions and detected loop-closure distributions.}
\label{fig:planning_vis}
\end{figure}

\begin{figure}[!t]
\centering
\includegraphics[width=\columnwidth]{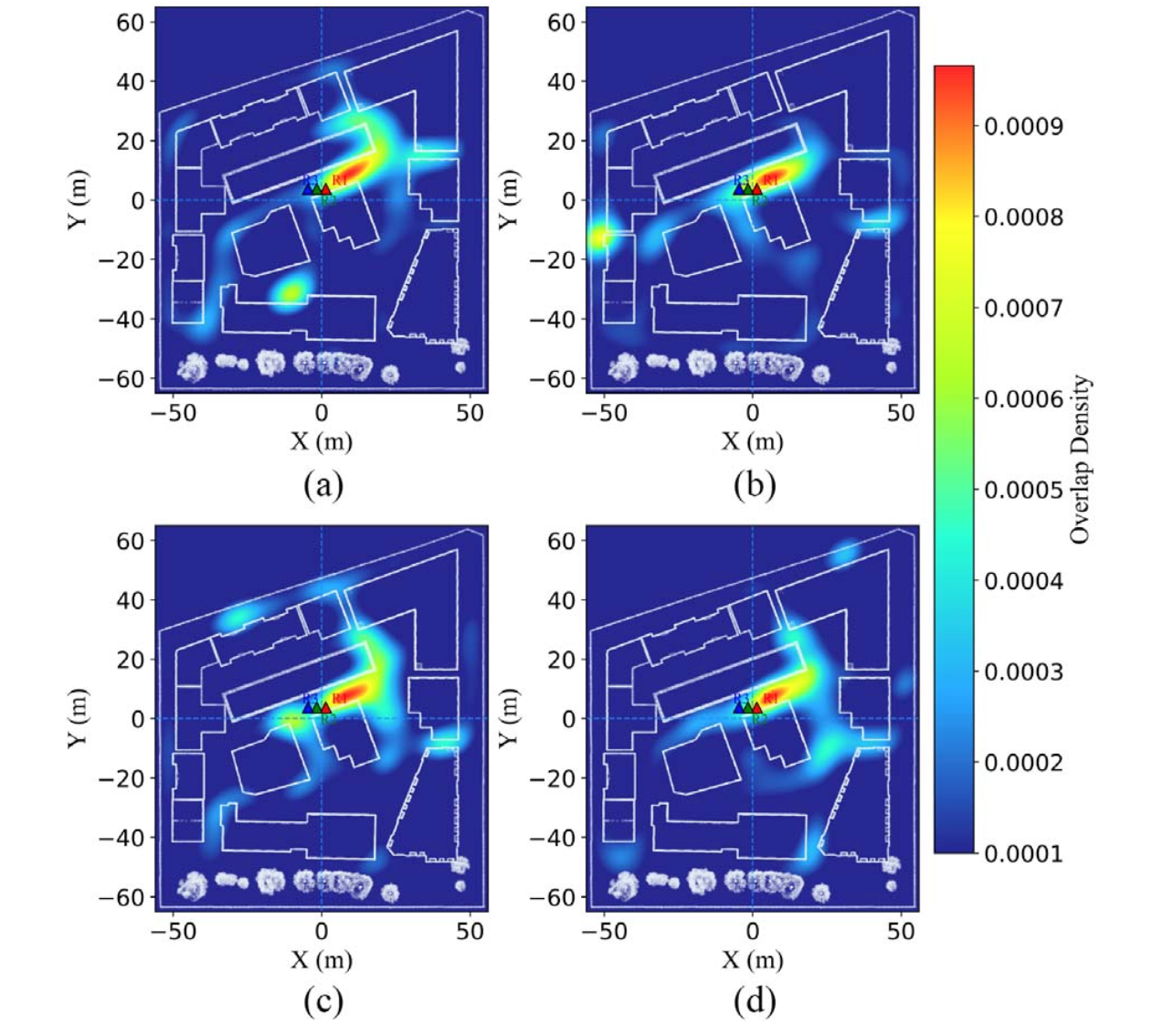}
\caption{Heatmap comparison of path overlap under different planning strategies. Brighter regions indicate higher trajectory overlap. The background point clouds depict the environment structure.}
\label{fig:planning_overlap}
\end{figure}

\subsection{System Evaluation and Communication Analysis}
End-to-end system performance is evaluated by combining different localization and planning modules in a structured campus environment with three UGVs.

\subsubsection{Impact of Exploration Strategy on Localization Accuracy}
Table~\ref{tab:comb_ate} compares Fast-LIO2, DCL-SLAM, and the proposed localization module under manual exploration, mTARE, and the proposed hierarchical planner. More informative and better-balanced trajectories lead to more useful loop closures and lower ATE. Under the proposed exploration strategy, all localization methods improve, and the proposed localization module consistently performs best.

\begin{table}[!t]
\caption{ATE (m) Under Different Combinations of Exploration and Localization}
\label{tab:comb_ate}
\centering
\resizebox{\columnwidth}{!}{%
\begin{tabular}{llccc}
\toprule
Exploration & Localization & UGV1 & UGV2 & UGV3 \\
\midrule
\multirow{3}{*}{Manual}
& Fast-LIO2~\cite{ref29} & 1.783 & 1.515 & 1.460 \\
& DCL-SLAM~\cite{ref8}   & 1.951 & 2.243 & 2.803 \\
& Ours (Loc.)            & 1.647 & 1.283 & 1.418 \\
\multirow{3}{*}{mTARE~\cite{ref1}}
& Fast-LIO2~\cite{ref29} & 1.289 & 2.415 & 2.260 \\
& DCL-SLAM~\cite{ref8}   & 1.248 & 2.363 & 2.041 \\
& Ours (Loc.)            & 1.203 & 1.971 & 1.785 \\
\multirow{3}{*}{Ours (Exp.)}
& Fast-LIO2~\cite{ref29} & 1.757 & 1.850 & 1.577 \\
& DCL-SLAM~\cite{ref8}   & 1.503 & 1.752 & 1.722 \\
& Ours (Loc.)            & 1.263 & 1.250 & 1.320 \\
\bottomrule
\end{tabular}}
\end{table}

\subsubsection{Impact of Localization Accuracy on Exploration Performance}
Table~\ref{tab:explore_metrics} shows exploration time, travel distance, area coverage, and path-balance index under two planners and three localization back ends. The proposed hierarchical planner consistently yields stronger area coverage and better path balance. When paired with the proposed localization module, it reaches full coverage with the shortest total time among the compared combinations.

\begin{figure}[!t]
\centering
\includegraphics[width=\columnwidth]{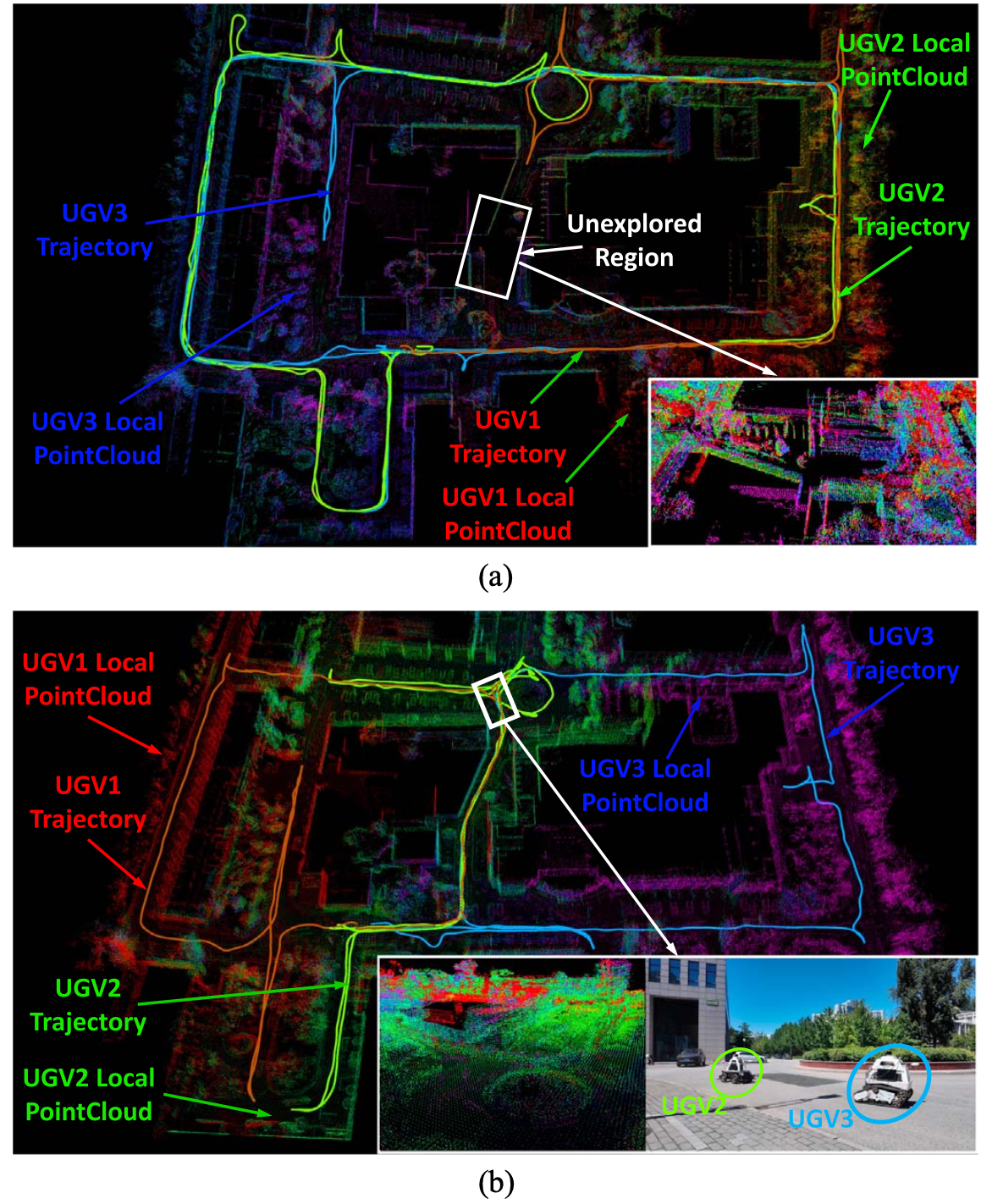}
\caption{Visual comparison of exploration results. Red, green, and blue lines indicate the trajectories of three UGVs. (a) Result using the mTARE planner with the proposed localization. (b) Result using the proposed planner with the proposed localization.}
\label{fig:exploration_visual}
\end{figure}

\begin{table*}[!t]
\caption{Exploration Metrics Under Different Localization Methods and Planning Strategies}
\label{tab:explore_metrics}
\centering
\resizebox{\textwidth}{!}{%
\begin{tabular}{lcccccc}
\toprule
\multirow{2}{*}{Metric} & \multicolumn{3}{c}{mTARE Planner~\cite{ref1}} & \multicolumn{3}{c}{Ours (Hierarchical) Planner} \\
& Fast-LIO2 & DCL-SLAM & Ours (Loc.) & Fast-LIO2 & DCL-SLAM & Ours (Loc.) \\
\midrule
Exploration Time (s)   & 300.24 & 402.35 & 432.25 & 337.43 & 349.56 & 331.18 \\
Travel Distance (m)    & 1611.83 & 2207.53 & 2412.48 & 1927.47 & 2012.31 & 1915.90 \\
Area Coverage (\%)     & 68.34 & 90.44 & 93.12 & 99.91 & 98.34 & 100.00 \\
Path Balance Index (\%)& 96.60 & 89.13 & 98.63 & 98.76 & 98.00 & 98.81 \\
\bottomrule
\end{tabular}}
\end{table*}

\subsubsection{Communication Overhead}
Table~\ref{tab:comm} compares message-type communication volume between the proposed framework and DCL-SLAM combined with the same exploration planner. The proposed framework requires much less bandwidth, mainly because dense keyframe point-cloud transfers are reduced and descriptor exchange remains compact.

\begin{table}[!t]
\caption{Communication Volume Per Message Type (MB)}
\label{tab:comm}
\centering
\resizebox{\columnwidth}{!}{%
\begin{tabular}{llcccc}
\toprule
Method & Type & UGV1 & UGV2 & UGV3 & Total \\
\midrule
\multirow{4}{*}{Ours}
& Odometry    & 1.52 & 0.62 & 0.78 & 2.92 \\
& Descriptors & 2.54 & 1.02 & 1.26 & 4.82 \\
& Keyframes   & 4.86 & 7.74 & 6.12 & 18.72 \\
& Topo.Map    & 9.60 & 6.58 & 9.34 & 25.52 \\
\midrule
\multirow{4}{*}{DCL-SLAM}
& Pose estimate & 0.60 & 0.88 & 0.76 & 2.24 \\
& Descriptors   & 12.06 & 13.32 & 12.88 & 38.26 \\
& Keyframes     & 40.99 & 72.40 & 71.12 & 184.51 \\
& Topo.Map      & 8.36 & 6.50 & 10.08 & 24.94 \\
\bottomrule
\end{tabular}}
\end{table}

\section{Conclusion and Future Work}
We present a fully distributed multi-UGV exploration framework for resource-limited environments with GPS denial, limited bandwidth, and the absence of prior maps. The framework unifies descriptor-aided localization, incremental topological mapping, and loop-aware hierarchical planning. A lightweight, viewpoint-invariant LiDAR descriptor with spectral-guided range-image alignment enables robust cross-UGV loop retrieval under substantial pose variations. Together with distributed pose-graph optimization, it enhances localization accuracy while maintaining global consistency.

To minimize communication overhead, only descriptor vectors and incremental subgraph updates are exchanged. For planning, a global MDVRP allocator coupled with a loop-aware local planner shortens exploration time and travel distance, supports informative loop-closure selection, and mitigates redundant path overlap. By coupling descriptor-aided localization with loop-aware hierarchical planning, the framework achieves both robust localization and efficient exploration.

Future research will focus on extending the framework to collaborative perception and planning between unmanned aerial vehicles and UGVs, enabling cross-platform loop closure detection, map fusion, and task-level coordination.

\end{document}